\documentclass[11pt,a4paper]{article}
\usepackage[hyperref]{acl2023}
\usepackage{times}
\usepackage{latexsym}

\usepackage{microtype}

\usepackage{paralist}
\usepackage{multirow}
\usepackage{amsmath}
\usepackage{graphicx}
\usepackage{csquotes}
\usepackage{float}

\usepackage{fancyhdr} 
\title{SpeakGer: A meta-data enriched speech corpus of German state and federal parliaments}

\author{Kai-Robin Lange \and Carsten Jentsch\\
Department of Statistics, TU Dortmund University, 44221 Dortmund, Germany\\
\texttt{\{kalange, jentsch\} @statistik.tu-dortmund.de}\\}

\begin{document}
\maketitle
\thispagestyle{fancy} 
\begin{abstract}
The application of natural language processing on political texts as well as speeches has become increasingly relevant in political sciences due to the ability to analyze large text corpora which cannot be read by a single person. But such text corpora often lack critical meta information, detailing for instance the party, age or constituency of the speaker, that can be used to provide an analysis tailored to more fine-grained research questions. To enable researchers to answer such questions with quantitative approaches such as natural language processing, we provide the SpeakGer data set, consisting of German parliament debates from all 16 federal states of Germany as well as the German Bundestag from 1947-2023, split into a total of 10,806,105 speeches. This data set includes rich meta data in form of information on both reactions from the audience towards the speech as well as information about the speaker's party, their age, their constituency and their party's political alignment, which enables a deeper analysis. We further provide three exploratory analyses, detailing topic shares of different parties throughout time, a descriptive analysis of the development of the age of an average speaker as well as a sentiment analysis of speeches of different parties with regards to the COVID-19 pandemic.
\end{abstract}

\section{Introduction}
In February of 2022, Germany's chancellor Scholz held a speech in the German Bundestag regarding the outbreak of the Russian-Ukrainian war of 2022. It was one of the most prolific speeches in a German parliament in the latest years due to its impact on Germany's foreign and defense policy, as it can be seen as the starting point for an increase in military spending and distancing towards the Russian government. But such decisions and speeches portrait the political stance of the speaker but not necessarily of the entire government or the speaker's party. We propose a data set with parliamentary debates from 16 German federal state parliaments as well as the German Bundestag over the time span of 76 years which is split into individual speeches with meta data to identify the current speaker. This meta data enables the analysis of topics, opinions and speech patterns of different politicians by party, political alignment, age, or constituency. We additionally identified comments from the audience, interrupting the speeches, to enable the analysis of crowd reactions to specific topics or speech patterns. We also labeled the speeches of session chairs: analyses can thus reduce the text corpus to only politically relevant speeches. As our data contains speeches from all 16 federal state parliaments, it can also be used to compare speeches across states to verify regional differences. We will publish the data set upon publication of this paper.

Further, we conduct an exploratory data analysis on the given corpus, using the ``party"{} meta data to analyze party topic shares as well as the sentiment of the 7 Bundestag parties in COVID-19 related speeches. We then use the ``age"{} indicator to analyze the developement of the average speaker age across time.
\section{Related Work}
In recent years, the interest in researching German political speeches by the means of Natural Language Processing has greatly increased. For instance, \citet{zeitenwenden} identify important political change points in the German political discourse using RollingLDA \cite{rieger.rolling, text2story}, a time-varying version of the topic model LDA \cite{LDA}, on a similar political data set of speeches of the German Bundestag. Another common research topic is the comparison of party positions \cite{ceronOptimizingTextRepresentations2022}, estimation of political alignment or ideological clarity of German and European political parties by using document scaling techniques. Some follow a classical bag-of-word approach \cite{PRR1, PRR2, wordfish, wordfish_usage, ideology_manifestos}, while others use topic models such as Top2Vec \cite{top2vec} to scale the available speeches or party manifestos \cite{TopicShoal}. Such analyses have also been extended to the predecessors of the Federal Republic of Germany, as \citet{DeuPARL} analyze political biases throughout the years using Reichstag as well as Bundestag data by using diachronic embeddings. Recent developments have also demonstrated the importance of claims and frames for the analysis of party positions, as \citet{blokkerWhyJustificationsClaims} exemplified using a data set of party manifestos. This aforementioned research does however often focus on the federal political level but disregards politics on the state level and below. And even at state level such analyses can often only differentiate their findings by party by using party manifestos over parliamentary speeches, as the available data sets used do not provide the necessary meta data. \citet{britishpolarity} also argues that such meta information is important to, for instance, measure political polarity in a supervised manner. The SpeakGer data set is meant to enable such fine-grained political research by meta-data enrichment.

In recent years, several similar data sets have been released which however lack some properties that are needed for quantitative text analysis of German parliaments. For instance, \href{https://openparliament.tv/}{Open Parliament TV} provide an interface for qualitative researchers for speeches in the German Bundestag from 2013 to 2023, split into individual speeches. This data set does however lack the speeches from the federal state parliaments and all Bundestag speeches prior to 2013. The ParlSpeech data set \cite{Parlspeech} provides split speeches of the German Bundestag from 1991 to 2018, but does not include speeches prior to this or from the 16 state parliaments. Still, \citet{Parlspeech} include information, to which agenda item the current speech refers to, which our corpus does not as of the publication of this paper, due to the different agenda and document structures across the 17 parliaments and the differences in stenographic reporting across 76 years. \citet{GerParCor} provide a similar data set which also includes parliamentary documents of the German Bundestag and the German federal state parliaments. This data set is also only provided in already pre-processed and part-of-speech-annotated form, while we publish unprocessed data to enable all researchers to apply pre-processing of their liking. We also split our data set by speeches and equip it with meta data about all speakers to enable a more fine-grained political analysis which includes meta-data such as the constituency, the party and the year of birth of all speakers while also allowing users to filter out speeches e.g. by session chairs and comments from the audience. Additionally, our data set contains data of the first 10 legislative periods of the federal state parliament of Berlin and the first 8 legislative periods of the federal state parliament of Baden-Württemberg.

\section{Data collection}
\label{Data}
We primarily recieved our data from the websites of the respective parliaments. However, some parliaments do not publish the documents of all legislative periods on their website, even if they are available. Thus, we collected additional documents from the \href{https://www.parlamentsspiegel.de}{Parlamentsspiegel-website} and looked for additional digitized documents in corresponding local museums. Still, not every legislative period of every German federal state parliament is digitized, as Bremen, Hamburg and Niedersachsen are missing digitized versions of the first legislative periods. However, representatives of all three federal state parliaments assured us that the remaining protocols are planned to be digitized as a part of a retro-digitization project. We therefore aim to update our data set as soon as the missing protocols are available to us. The source of each protocol gathered is detailed in \autoref{tab:sources}. To enable a time-dependent analysis, we collected the exact dates for each plenary session of all 17 parliaments and integrated these dates into our meta data. We directly received this information from the respective parliament officials we contacted.

\begin{table*}[h]
    \centering
    \caption{Sources and links to all protocols that were analyzed. If the protocols of a parliament cannot be found in one place, we provide multiple sources for all possible legislative periods.}
    \begin{tabular}{|c|c|c|}
    \hline 
    \multirow{2}{*}{Parliament (English name)}& Legislative & \multirow{2}{*}{Source}\\
    & period & \\
    \hline\hline
    Baden-Württemberg & 12-17& \href{https://www.landtag-bw.de/home/dokumente/plenarprotokolle.html}{Landtag von Baden-Württemberg} \\
    (Baden-Wuerttemberg) & 1-11 &  \href{https://www.wlb-stuttgart.de/literatursuche/digitale-bibliothek/digitale-sammlungen/landtagsprotokolle/digitale-praesentation/zeitliche-gliederung/zeitraum-1952-1996/}{Württembergische Landesbibliothek}\\
    \hline
    Bayern (Baveria) & 1-18& \href{https://www.bayern.landtag.de/webangebot2/webangebot/protokolle?execution=e1s1}{Bayrischer Landtag} \\
    \hline
    \multirow{3}{*}{Berlin} & 12-19 & \href{https://pardok.parlament-berlin.de/portala/start.tt.html}{Abgeordnetenhaus Berlin}\\
    & 6-11 & \href{https://digital.zlb.de/viewer/metadata/15975513/1/}{Zentral- und Landesbibliothek Berlin}\\
    & 1-5 & \href{https://digital.zlb.de/viewer/metadata/15975510/0/}{Zentral- und Landesbibliothek Berlin}\\
    \hline
    \multirow{2}{*}{Brandenburg} & 8-10 & \href{https://www.landtag.brandenburg.de/de/parlament/plenum_und_gesetze/uebersicht_der_plenarsitzungen/25212}{Landtag Brandenburg}\\
    & 1-7 & \href{https://www.parlamentsspiegel.de/home/bestandsubersicht-drucksachen-un.html}{Parlamentsspiegel}\\
    \hline
    \multirow{2}{*}{Bremen} & 18-20 & \href{https://www.bremische-buergerschaft.de/index.php?id=570}{Bremische Bürgerschaft}\\
    & 7-17 & \href{https://www.parlamentsspiegel.de/home/bestandsubersicht-drucksachen-un.html}{Parlamentsspiegel}\\
    \hline
    Bundestag & 1-20 & \href{https://www.bundestag.de/services/opendata}{Deutscher Bundestag}\\
    \hline
    \multirow{2}{*}{Hamburg} & 20-22 & \href{https://www.buergerschaft-hh.de/parldok/formalkriterien/1}{Hamburgerische Bürgschaft}\\
    & 6-19 & \href{https://www.parlamentsspiegel.de/home/bestandsubersicht-drucksachen-un.html}{Parlamentsspiegel}\\
    \hline
    Hessen & 1-20 & \href{https://starweb.hessen.de/starweb/LIS/plenarprotokolle.htm}{Hessischer Landtag}\\
    \hline
    Mecklenburg-Vorpommern & \multirow{2}{*}{1-8} & \multirow{2}{*}{\href{https://www.landtag-mv.de/parlamentsdokumente}{Landtag Mecklenburg-Vorpommern}}\\
    (Mecklenburg-Western Pomerania) & & \\
    \hline
    \multirow{2}{*}{Niedersachsen (Lower Saxony)} & 17-18 & \href{https://www.landtag-niedersachsen.de/parlamentsdokumente/parlamentsdokumente/}{Landtag Niedersachsen}\\
    & 8-16 & \href{https://www.parlamentsspiegel.de/home/bestandsubersicht-drucksachen-un.html}{Parlamentsspiegel} \\
    \hline
    Nordrhein-Westfalen (North Rine Westfalia) & 1-18 & \href{https://www.landtag.nrw.de/home/dokumente/dokumentensuche.html}{Landtag Nordrhein-Westfalen}\\
    \hline
    Rheinland-Pfalz (Rhineland Palatinate)& 1-18 & \href{https://opal.rlp.de/starweb/OPAL_extern/servlet.starweb?path=OPAL_extern/SUCHEXPOR_TREE.web&search=R\%3D1841}{Landtag Rheinland-Pfalz}\\
    \hline 
    \multirow{2}{*}{Saarland} & 14-17 & \href{https://www.landtag-saar.de/dokumente?searchValue=&OnlyTitle=false&Categories=Print,PlenaryProtocol,Law,PublicConsultation,Operations&DateFrom=&DateTo=&periods=&SortValue=Erscheinungsdatum&SortOrder=desc&tab=Doc&DocumentPage=true}{Landtag des Saarlandes}\\
    & 7-13 & \href{https://www.parlamentsspiegel.de/home/bestandsubersicht-drucksachen-un.html}{Parlamentsspiegel} \\
    \hline
    Sachsen (Saxony) & 1-8 & \href{https://edas.landtag.sachsen.de/}{Sächsischer Landtag}\\
    \hline
    \multirow{2}{*}{Sachsen-Anhalt (Saxony-Anhalt)} & 6-8 & \href{https://www.landtag.sachsen-anhalt.de/dokumente/aktuelle-dokumente/plenarprotokolle}{Landtag von Sachsen-Anhalt}\\
    & 1-5 &  \href{https://www.parlamentsspiegel.de/home/bestandsubersicht-drucksachen-un.html}{Parlamentsspiegel} \\
    \hline
    Schleswig-Holstein & 1-20 & \href{http://lissh.lvn.parlanet.de/shlt/start.html}{Schleswig-Holsteiner Landtag}\\
    \hline
    \multirow{2}{*}{Thüringen (Thuringia)} & 4-7 & \href{https://www.thueringer-landtag.de/plenum/protokolle/}{Thüringer Landtag}\\
    & 1-4 & \href{https://www.parlamentsspiegel.de/home/bestandsubersicht-drucksachen-un.html}{Parlamentsspiegel} \\
    \hline
    \end{tabular}
    \label{tab:sources}
\end{table*}

\subsection{Text extraction and spelling correction}
Out of the available 240 legislative periods, the protocols of a total of 106 periods are either available as text files or pdf-files from which text can be extracted. Some  of the remaining documents are scanned pdf files in which each page of the protocol is only displayed as a picture with no possibility for direct text extraction. To extract the text from these documents, we use Google's tesseract \cite{tesseract}, a model for Optimal Character Recognition (OCR), with the German language option (and a Fraktur-option for the first legislative period of the state Bayern). We improve tesseract's performance by binarizing each page to a pure black-white format using Otsu's threshold \cite{otsu} and by correcting a possible skew of each page using OpenCV \cite{OpenCV}. We found tesseract to best capture the text of two-column documents in a sample of our data we used as an experiment. 

Such an OCR model is however not able to detect a text perfectly, but will, especially for older and less clean fonts, yield ``spelling"{}-errors. That is, despite not literally spelling the word, single letters of a word can be misinterpreted as a different letter, having a similar effect to a misspelled word. The term ``Bravo!"{} is for instance often misclassified as ``Bravol"{} by tesseract. We contemplated using a prediction-based spelling correction, e.g.~a masked word prediction based on BERT \cite{BERT}, but due to frequent mistakes in particularly old documents, this context-based prediction yielded sub-optimal results. To correct the errors that are caused by such OCR models, we therefore aim to instead use a lexicon-based approach by using Symspell's \cite{Symspell} German language dictionary which we additionally provided with the last names of all members of parliament (mps) of all 16 federal state parliaments to stop the spelling correction from affecting our speech-splitting. We detect every word in every OCR scanned document that is not part of this dictionary and determine, whether there is a word in the dictionary that is sufficiently similar to the misspelled word with regards to their Levenshtein-distance \cite{levenshtein}, that is the number of character transformations needed to turn one misspelled token into a correctly spelled token. This distance is chosen dynamically, depending on the word's length. For instance, a word with 7 characters is allowed to have a larger levenshtein-distance to it's ``correct spelling"{} than a word with just two characters. We publish both the spell-checked versions as well as the original processed documents.

\subsection{Speech splitting}
To identify speeches, we first gathered crucial information about possible speakers by scraping meta data about the first name, last name, year of birth, party, constituency and Wikipedia-links of each speaker, if available. For this, we used the Wikipedia-pages of each federal state parliament, detailing all participating members of parliament during each legislative period. To simplify the interpretation of smaller and regional parties, we also include the political alignment of the parties according to their Wikipedia-pages (e.g. left-populist, social democratic, liberal or conservative). The regular expressions used to identify the start and end of a plenary session as well as splitting the speeches can be found in our \href{https://github.com/K-RLange/SpeakGer}{GitHub-repository}. We will also use said GitHub-repository to detail link and update on the publication of the data set. In the following paragraphs, we describe how they are designed as well as their purpose.

To split the speeches, we first determine, where the plenary session starts and when it ends to cut off the table of contents and a possible appendix to the pdf-file. To account for possible OCR mistakes, we use Regular Expressions to identify either a comment such as ``(Beginn: ... Uhr)" marking the start of a session, or, if this cannot be detected, the first appearance of common speech patterns, such as a greeting like ``Meine sehr verehrten Damen und Herren". We also incorporate common OCR errors for those phrases in our Regular Expressions, such as misinterpreting an ``B"{} as an ``ß"{}. To find the end of the session, we look for either a comment marking the end of the session similar to ``(Ende: ... Uhr)" or we end the session when we detect common speech patterns, which are used to close a session like ``die Sitzung ist damit geschlossen" or ``Ich schließe damit die Sitzung". If none such indicators are found, which usually only happens in old documents with bad quality scans, we heuristically cut the last/first 1000 lines of our document to remove the table of contents and appendix. 

After detecting in which part of the document the speeches take place, we split the remaining text into pieces with the use of Regular Expressions and our meta data. All documents have common styles which can be used to identify comments and the start of a speech.

Speeches can be identified by a string search for each line by looking for the last name of said mp, followed by a colon. There are some variations of this, such as including the word ``Abgeordneter" or a title before stating the name (``Abgeordneter Dr. Mustermann:"), or the party of the mp (``Mustermann (SPD):"), but the last name of the mp as well as the colon are always present across all analyzed parliaments. Thus, we detect a change in speakers by scanning the lines for the last names of all possible mps in this legislative period paired with a colon. For this we use the names from the mps of the parliament and legislative period that are analyzed, which were scraped from Wikipedia. If we detect the word ``Präsident"{} or get another indication that the speech is held by the chair of the session, we mark it accordingly, as it will likely only cover the organization of the plenary session and rarely contains political statements or arguments.

As a comment, we define additional information provided by the stenographer about the organization of the session (such as information on pauses when the parliament votes on a bill) as well as interjections from the audience during a speech. Such comments can be identified, as they are surrounded by either square or round brackets. Some contain an interjection from a specific member of the parliament, which is detected if the last name of an mp is used in the comment, or about reactions of certain parties, which are detected if said party names are used in the comment. Otherwise, the meta data regarding the speaker is set to ``unknown" for such comments. We consider a speech that is interrupted by such a comment to be two separate speeches, before and after the comment, held by the same speaker. This is done to enable the analysis of interactions between comments and speeches such that the effect of a comment on the speech or vice versa can be analyzed.

\section{Descriptive Analysis}
In total, the SpeakGer data set contains 17,784,802 texts across the 16 German federal state parliaments as well as the German Bundestag, which include a total of 5,510,951 comments, 1,467,746 speeches of session chairs and 10,806,105 speeches of other mps. The total number of documents (in thousands), split into comments, speeches of the session chair and other speeches, separated by parliament are displayed in \autoref{tab:complete_data}. 
\begin{table*}[t]
\caption{Total number of speeches in thousands in each parliament, divided by party of speaker and whether the speech is a comment or given by the chair of the session. As the party Die Linke is the successor of the parties SED and PDS, we look at the speeches of said parties combined.}
\begin{tabular}{lccccccccc}
\hline
Parliament &   Chair &  Comment &  AfD &  CDU &  CSU &  FDP &  Grünen &  SPD & Linke\\
\hline
Bundestag                    &    378 &          1168 &   26 &  644 &  144 &  282 &         167 &  605 &            116 \\
Baden-Württemberg      &     18 &           555 &   14 &  319 &    0 &  106 &          97 &  238 &              0 \\
Bayern                 &    110 &           312 &    2 &    0 &  263 &   14 &          30 &  115 &              0 \\
Berlin                 &     38 &           263 &   13 &  144 &    0 &   48 &          49 &  172 &             46 \\
Brandenburg            &     39 &            75 &    9 &   34 &    0 &    4 &           9 &   83 &             37 \\
Bremen                 &     37 &           254 &    0 &   91 &    0 &   29 &          33 &  162 &              9 \\
Hamburg                &     69 &           337 &    4 &  120 &    0 &   32 &           7 &  168 &             14 \\
Hessen                 &    110 &           390 &    7 &  247 &    0 &   74 &          95 &  220 &             24 \\
Mecklenburg-Vorpommern &     55 &           334 &   22 &  120 &    0 &   15 &          11 &  119 &            113 \\
Niedersachsen          &    109 &           226 &    0 &  181 &    0 &   57 &          68 &  152 &              7 \\
Nordrhein-Westfalen    &    133 &           503 &    8 &  201 &    0 &   73 &          72 &  271 &              5 \\
Rheinland-Pfalz        &     56 &           253 &    7 &  110 &    0 &   29 &          24 &  132 &              0 \\
Saarland               &     34 &           122 &    1 &   78 &    0 &    9 &           9 &   76 &              6 \\
Sachsen                &     67 &           129 &   11 &   97 &    0 &   11 &          18 &   35 &             50 \\
Sachsen-Anhalt         &     49 &           106 &   14 &   67 &    0 &   10 &          17 &   32 &             35 \\
Schleswig-Holstein     &     87 &           327 &    0 &  157 &    0 &   68 &          29 &  170 &              2 \\
Thüringen              &     62 &           152 &    9 &   69 &    0 &   11 &           8 &   28 &             41 \\

\hline
\end{tabular}
\label{tab:complete_data}
\end{table*}
\subsection{Topic shares per party}
To determine topic shares per party over time, we use RollingLDA \cite{rieger.rolling}, a rolling window approach to topic modeling that creates coherently interpretable topics modeled over time that are allowed to adapt to a changing vocabulary. We thus receive a topic model each year from 1950 to 2022. The years 1947 to 1950 are used to fit the initial model while later years update the model that came beforehand. For this, we consider $K=30$ topics to give the topic model the opportunity to separate a wide range of political aspects in different topics but still enabling a clear analysis in the scope of this paper. We additionally set the parameters $\alpha=\gamma=\frac{1}{K}$ and the memory-parameter to $4$, thus enabling the model to ``remember"{} the previous 4 years to create topics in the current year. We fit our model on the data of all federal state parliaments simultaneously but only use speeches that were not classified as comments to prevent topics simply representing crowd reactions like applause. 

\begin{figure*}[t]
\centering
\includegraphics[width=\textwidth]{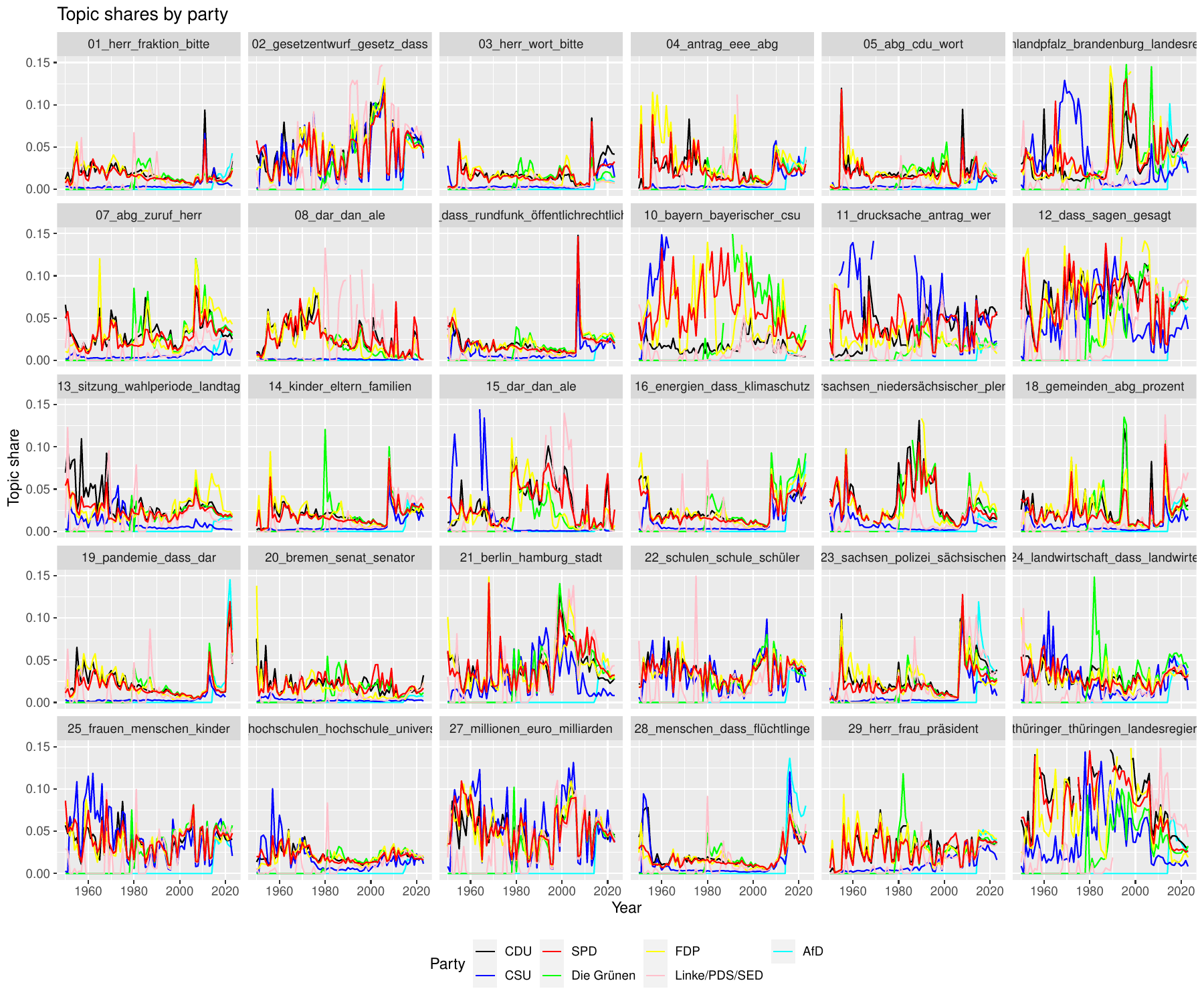}
\caption{Topic shares of the current 7 Bundestag-parties from 1950 to 2022 in the 16 German federal state parliaments. As the party Die Linke is the successor of the parties SED and PDS, we look at the speeches of said parties combined.}
\label{fig:RollRes}
\end{figure*}

The topic shares for each topic over time, separated by party are displayed in \autoref{fig:RollRes}. For this figure, we used the ggplot-package \cite{ggplot} for the R programming language \cite{R}. For better visibility, we limited the plots to topic shares up to 15\%, which only has minor implications for most topic. Only the topic share of the Baverian party CSU is off the charts for most of topic 10 and 11, as these cover topics extensively covered in the Baverian parliament. In the figure, the topics' top words over the entire time period are used to title the respective topics graphs. These overall top words most often are not the top words at all times, but still decently represent said topic as a whole. Speeches that are part of documents with particular bad scan quality often contain a lot of misspelled words, which leads to topics that are characterized by commonly misspelled words -- this can be seen by observing the top words ``dar"{}, ``dan"{}, ``ale"{} (which are likely misspelled versions of the words ``das"{} and ``alle"{}) of the topics 8 and 15. This filtering aspect of the topic model allows us to focus on the other, relevant topics without the need to account for misspelled words - also due to the properties of RollingLDA, these topics ``rotate out"{} as soon as the OCR errors disappear.

Due to the fact that we perform a topic analysis on documents from all 16 federal state parliaments, several German states have a specific topic designated to them, which can be interpreted as the talk about local affairs. Despite talking about different places, some of these topics overlap, possibly due to similar actions that need to be taken -- the city states Berlin, Bremen and Hamburg have a joint topic dealing with city state affairs (Topic 18). We can further inspect the respective topics of these states to gather information about the most important discussion in said parliament at the time. To further analyze the contents of each parliament though, a detailed topic analysis can be performed on only those documents that belong to said parliament. Apart from these topics, which specifically define misspelled words or German states, topics 9, 14, 16, 19, 22, 24, 25, 26, 27 and 28 also cover more general topics that are of interest in every federal state, for instance education, climate change and state-finances. The rest of the topics cover parliament-specific vocabulary like ``drucksachen"{} or ``gesetzesentwurf"{} in topics 11 and 2 respectively.

The topics of political interest confirm several political assumptions to parties that can be made by observing the parties in the Bundestag and considering their party manifestos on a federal level. For instance, we can observe the green party Die Grünen, having the highest topic shares of all parties in topics 16 and 24 covering climate change and agriculture respectively. The party CSU that is only present in the federal state Bavaria, which contains a lot of rural areas, also talks a lot about agriculture while talking the least of all parties about renewable energies and climate change. Conversely, the liberal party FDP, whose party manifestos focus on new technology, have a high topic share in the topic about climate change and renewable energy, while barely talking about agriculture.

For the right-wing party AfD, we observe a high topic share in the topics 19 and 28. Starting from 2020, topic 19 covers the COVID-19 pandemic during which the AfD was very vocal about opposing the lockdowns and other restrictions of the government to prevent the spread of the virus. Topic 28 covers the refugee crisis in Germany starting in 2014, which has been one of the AfD's biggest topics since it was founded in 2014. In 2022, topic 28 transformed about a topic about the Russian-Ukrainian war with major parts of the major German parties AfD and Die Linke supporting Russia in the conflict. This is also reflected in our topic models, as both these parties have the highest share of all parties in this topics.

Interestingly, the two biggest parties of Germany, the SPD and CDU barely dominate the shares in any topic. This is likely because these two parties are considered the most centrist parties, that cover a broad range of political topics without extensively focusing on a specific topic.

Overall, the behavior of the major 7 German parties on the state level reflects their behavior on the federal level in the Bundestag. This analysis however only demonstrates this while looking at all federal states combined, to investigate whether this applies only ``on average"{} or in all parliaments, said parliaments need to be evaluated individually.

\subsection{Sentiment Analysis}
As a further descriptive analysis of our data set, we perform a party-based sentiment analysis across each parliament to see if any party's speeches are particularly positive or negative in speeches regarding the COVID-19 pandemic. As there is no training data set available, we perform an unsupervised sentiment analysis. For this we use Lex2Sent \cite{l2s}, an unsupervised sentiment analysis tool that uses Doc2Vec \cite{d2v} to enhance a lexicon-based sentiment analysis. This approach allows us to specify, how a positive or negative sentiment can be determined for political speeches compared to regular web documents as it is based on a sentiment lexicon specifically catered for this task. Lex2Sent further improves the classical lexicon-approach by measuring the distance of a document to both the positive and negative half of a lexicon using Doc2Vec, which is trained on resampled documents of the original corpus. This resampling leads to a bagging-effect which boosts the performance of this analysis. To enable a political analysis using Lex2Sent, we use the sentiment dictionary for German political language as a lexicon-base for Lex2Sent \cite{rauh}. 

\begin{figure}[h]
\centering
\includegraphics[width=\linewidth]{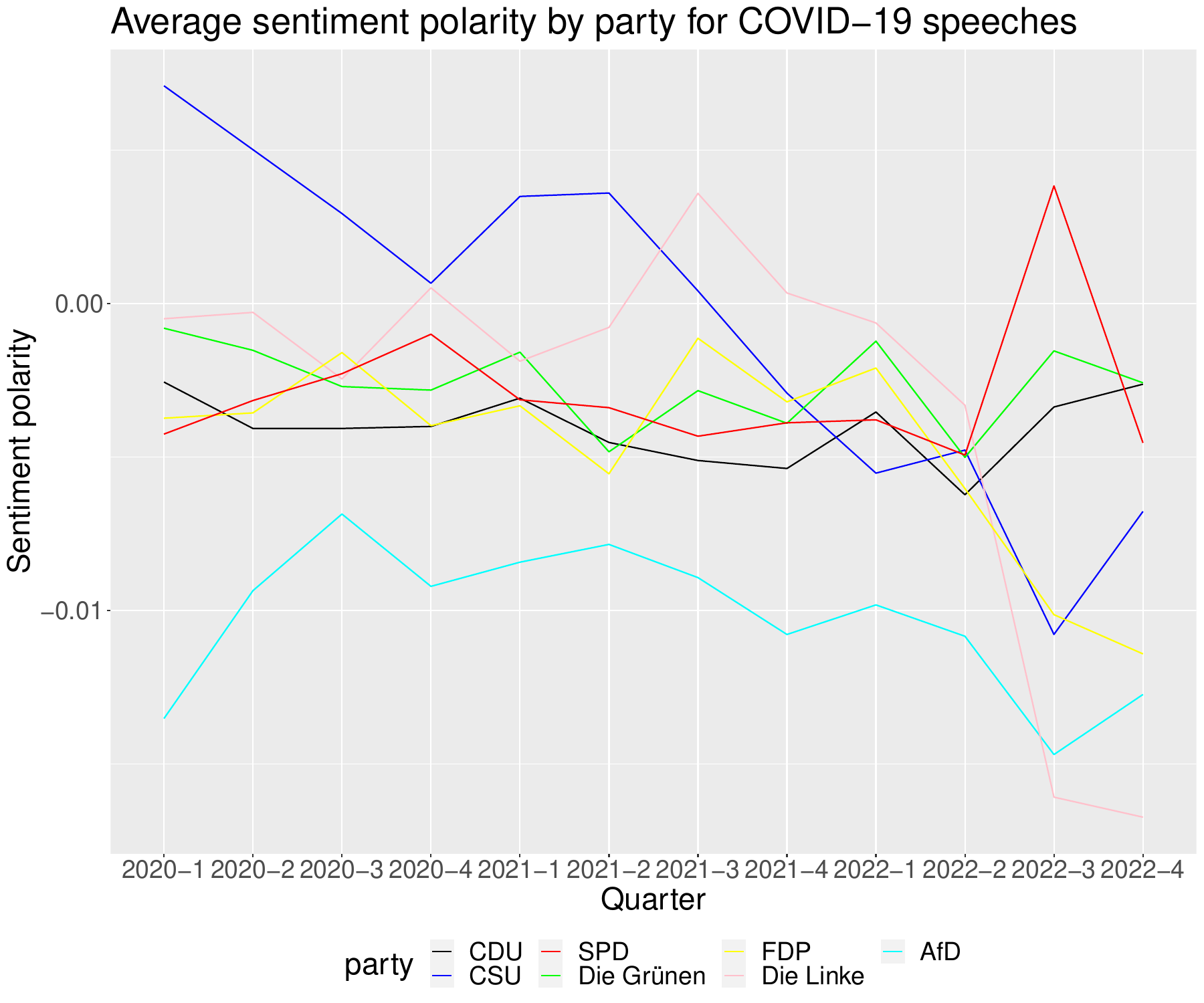}
\caption{Average sentiment polarites of COVID-19 related speeches of the 7 Bundestag parties in all 17 state and federal parliaments of Germany from 2020 to 2022. The scores were calculated using Lex2Sent, where a negative value indicates a negative speech.}
\label{fig:l2sRes}
\end{figure}

In \autoref{fig:l2sRes}, we display the average sentiment polarity, calculated by Lex2Sent, for each party in 2020 to 2022. The larger the sentiment polarity, the more positive a speech is estimated, with negative values indicating rather negative speeches. We can see that the average sentiment across all parties is rather negative, which is not surprising given the topic at hand. Terms such as ``Pandemie"{} are generally considered to be negative and the speeches thus generally have a negative undertone. What is more interesting is the comparison of the parties' sentiment. For instance, speeches of the right-wing party AfD, which heavily protested the COVID-19 lockdowns and restrictions, are considered to be the most negative in ten of the twelve observed quarters by the model.

In the last two quarters of 2022, the left-wing party Die Linke shows a more negative average sentiment compared to all other parties including the AfD. This is despite them generally delivering positive speeches until this point. One reason for this might be change of party doctrine following Russia's invasion of Ukraine. As mentioned before, major parts of Die Linke are considered to be Russian-favored. The debates resulting from the war outbreak might have thus caused the party to become more confrontational with other parties as a whole. This explanation should however be taken with caution, as the number of speeches concerning COVID-19 has greatly decreased in the last two quarters of 2022 and the observed negative sentiment could this be result of this low sample size.

The Bavarian party CSU also shifted its sentiment over time. As seen in \autoref{fig:l2sRes}, the CSU starts off, having the most positive average sentiment in their speeches concerning COVID-19. During this time, the CSU were party of the government in both the Bundestag and the Bavarian state parliament. During this time the party, and especially their party leader Markus Söder, advocated in favor of hard lockdowns and restrictions. The CSU was thus very in-line with actions taken by the government to handle the pandemic. We see a shift in sentiment starting during the election campaign in the third quarter of 2021, worsening after the elections in 2021. This might be result of the CSU itself not being part of the German federal government anymore and thus not being so compliant with the actions of the government any more.

The same cannot be said for the CSU's sister party CDU however, as the conservative party's sentiment remains rather average across time. The same goes for Die Grüne, the FDP and the SPD.
\subsection{The age of speakers}
\begin{figure}[h]
\hspace{-0.3cm}
\includegraphics[width=1.1\linewidth]{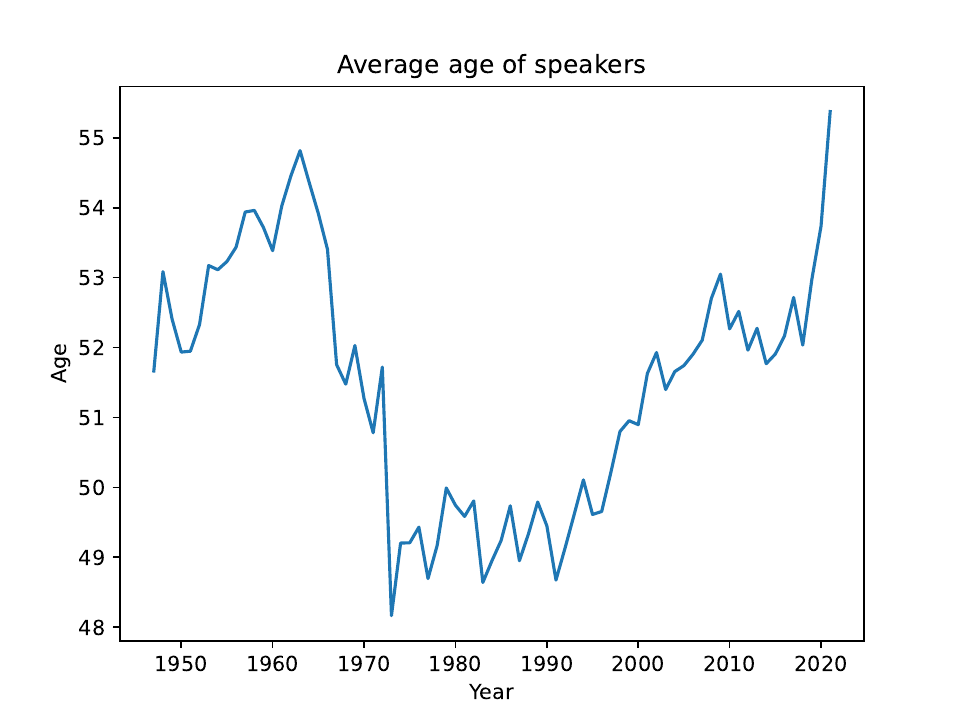}
\caption{Average age of speakers in German parliaments from 1947 to 2022.}
\label{fig:age}
\end{figure}

In this subsection, we focus on the age of the speakers across Germany. The average age of all registered speakers in the SpeakGer data set from 1947 to 2022 is displayed in \autoref{fig:age}. We can see that the average age of speakers started to decrease from 54.82 in 1963 to 48.17 in 1973. While the average age remained similar until 1991, the average speaker age started increasing after the German Reunification in 1990. Ultimately, the average speaker age continued to increase, reaching its maximum of 55.38 in 2022. This is partly due to the increasing age of CDU-speakers. While speakers of Germany's largest conservative party averaged at 54.37 years of age in 2018, this increased to an average of 60.04 years in 2021.
\section{Summary}
We propose the SpeakGer corpus, a comprehensive text data set detailing the long history of German parliamentary debates across 16 federal state parliaments as well as the German Bundestag, split into statements of the session chair, comments and interjections as well as speeches of members of the parliament. Each individual speech is equipped with rich meta data, such as the date of the speech, the party of the speaker and the political alignment of said party, the speaker's age and the speaker's constituency. In total, the SpeakGer data set contains 10,806,105 speeches. This enables researchers to perform fine-grained political analyses of the data set, in which different parties, age-groups and states can be compared. As an exemplary usage of the data set, we performed unsupervised sentiment analysis as well as time-dependent topic modeling to our data and demonstrate how even simple analyses can provide interesting results with the help of meta data. Our results indicate that regional alterations of Bundestag parties often follow the lead of the federal party, despite regional differences, as the sentiment and topics align with the behavior of the parties on a federal level. For instance, the left-wing party Die Linke appears to follow a more confrontational approach to speeches in federal state parliaments after the outbreak of the Russian-Ukrainian war, even in seemingly unrelated topics such as COVID-19 and despite being part of regional state governments themselves. This is however only a preliminary result of our exploratory analysis and should be inspected further.

In future research, we aim to, among other possible research ideas, further use the SpeakGer data set proposed to inspect, validate and broaden our preliminary results on the differences between regional and federal versions of the same party. As we only focused on the ``party"{} information in our exploratory research in this paper, in future research, we intend to use the remaining meta data, such as the age of the speaker or the speaker's constituency to perform analyses that take spatial and regional aspects into account.

\section*{Ethical considerations}
We provide this data set with best intentions to enable researchers to gain a new perspective on German politics. We only use publicly available information to equip our corpus with meta data. We however cannot be certain that the data will not be misused to push political agendas by for instance framing a specific party. We do believe that the benefits of such a publicly available data set outweigh the possible negative aspects, as such malicious framing is commonly done without using a data set of federal state parliament speeches.

\section*{Limitations}%
As a result of sub-optimal document-scans in earlier legislative periods in almost all federal state parliaments, not all speeches and speakers could be correctly identified. In addition to this, old scans of the state Nordrhein-Westfalen contain not just one plenary session but multiple, which also had to be manually split. This session splitting might be sub-optimal due to the poor quality scans. While we contacted all federal state parliaments about the specific dates for all plenary sessions and most states were able to provide a complete list of all correct dates, the states Berlin, Niedersachsen and Schleswig-Holstein could only provide us with an incomplete list. Thanks to publicly available information on Wikipedia, we were able to estimate the dates for the missing plenary sessions of these states, which are however subject to some noise. Lastly, as a result of the meta-based splitting of speeches, we are not able to detect speeches of guests of the parliament, such as Wolodomyr Selensky speaking in the German Bundestag on March 17th 2023, as these guests' names are not part of our meta data containing only information about the mps of the parliament. We aim to improve on these aspects of the data set as soon as better OCR methods and the results of the retro-digitization project of the German federal state parliaments are released.
\section*{Acknowledgments}
This paper is part of a project of the Dortmund Center for data-based Media Analysis (\href{https://docma.tu-dortmund.de/}{DoCMA}) at TU Dortmund University. The work was supported by the Mercator Research Center Ruhr (MERCUR) with project number Pe-2019-0044. In addition, the authors gratefully acknowledge the computing time provided on the Linux HPC cluster at TU Dortmund University (LiDO3), partially funded in the course of the Large-Scale Equipment Initiative by the German Research Foundation (DFG) as project 271512359.

\bibliographystyle{acl_natbib}
\bibliography{acl_latex}

\end{document}